\documentclass{article}

\PassOptionsToPackage{square,numbers}{natbib}

\usepackage[preprint]{nips_2018}

\usepackage[utf8]{inputenc} 
\usepackage[T1]{fontenc}    
\usepackage{hyperref}       
\usepackage{url}            
\usepackage{booktabs}       
\usepackage{amsfonts}       
\usepackage{nicefrac}       
\usepackage{microtype}      

\usepackage{graphicx}
\usepackage{pifont}
\usepackage[linesnumbered,ruled]{algorithm2e}
\usepackage[scaled=0.8]{inconsolata}

\title{Bayesian Inference of Regular Expressions from Human-Generated Example Strings}

\usepackage{color}
\definecolor{Blue}{RGB}{10,100,200}

\usepackage{listings}
\definecolor{gray}{rgb}{0.2,0.2,0.2}
\definecolor{red}{rgb}{0.8,0.3,0.3}
\definecolor{lightgray}{rgb}{.99,.99,.99}
\definecolor{darkgray}{rgb}{.4,.4,.4}
\definecolor{purple}{rgb}{0.65, 0.12, 0.82}
\definecolor{orange}{rgb}{1,0.5,0}

\newcommand{\posx}{\textcolor{Blue}{\ding{51}}}%
\newcommand{\negx}{\textcolor{red}{\ding{55}}}%
\newcommand{\exsep}{{\vline height 2ex}}
\newcommand{\exsepb}{{\vline height 2.5ex}}

\lstdefinelanguage{JavaScript}{
  keywords={typeof, new, true, false, catch, function, return, null, catch, switch, var, if, in, while, do, else, case, break},
  keywordstyle=\color{blue}\bfseries,
  ndkeywords={class, export, boolean, throw, implements, import, this},
  ndkeywordstyle=\color{darkgray}\bfseries,
  identifierstyle=\color{black},
  sensitive=false,
  comment=[l]{//},
  morecomment=[s]{/*}{*/},
  commentstyle=\color{purple}\ttfamily,
  stringstyle=\color{red}\ttfamily,
  morestring=[b]',
  morestring=[b]"
}

\usepackage{pifont}

\author{
  Long~Ouyang\thanks{Thanks for Leon Bergen, Andreas Stuhlmuller, and Tim O'Donnell for invaluable feedback} \\
  \texttt{longouyang@post.harvard.edu} \\
}

\begin{document}

\maketitle

\begin{abstract}
  In programming by example, users ``write'' programs by generating a small number of input-output examples and asking the computer to synthesize consistent programs.
  We consider a challenging problem in this domain: learning regular expressions (regexes) from positive and negative example strings.
  This problem is challenging, as (1) user-generated examples may not be informative enough to sufficiently constrain the hypothesis space, and (2) even if user-generated examples are in principle informative, there is still a massive search space to examine.
  We frame regex induction as the problem of inferring a probabilistic regular grammar and propose an efficient inference approach that uses a novel stochastic process recognition model.
  This model incrementally ``grows'' a grammar using positive examples as a scaffold.
  We show that this approach is competitive with human ability to learn regexes from examples.
\end{abstract}

\section{Introduction}

\setlength{\fboxsep}{2pt}

In many domains, people use discrete rules as a critical part of their workflow.
For example, computer users may wish to transform columns in a spreadsheet according to clearly defined functions \cite{gulwani11}, or extract certain elements from a webpage \cite{chasins15}.
In some cases, we would like to synthesize such rules from \emph{examples} like input-output pairs.
This is advantageous if the end user does not know how to program rules or has clear examples in mind but wishes to outsource the burden of writing the rule.
This is known as \emph{programming by example}.
This paper considers a challenging programming by example problem: learning string classifiers.
In particular, we learn a small but realistic subset of regular expressions (regexes) from human-generated positive and negative example strings.
Regular expressions are an attractive target for programming by example, as filtering strings using rules is a common task (e.g., text editing or performing input validation), but it can be quite difficult for laypeople to learn how to write regular expressions themselves and even experts can struggle with debugging them.

Inferring regular expressions from examples is challenging for two reasons.
First, the dataset of user-supplied examples may be uninformative.
For instance, consider the following dataset that was generated by a subject in the human experiments of \cite{ouyang17}:   \texttt{\framebox{dogs {\posx} {\vline height 2ex} cat {\negx}}} (i.e., a single positive string \texttt{dogs} and a single negative string \texttt{cat}).
Intuitively, there are many equally plausible hypotheses here---the string must be four or more characters, the string must begin with \texttt{d}, the string must contain an \texttt{o}, and so on.
As it happens, the intended regex that this subject was trying to convey was \texttt{.*s} (the string must end in an s), but the evidence provided does not support this hypothesis very well (or any other hypothesis, for that matter).
Second, even if the dataset is in principle informative about the intended regex, finding good hypotheses is computationally challenging---the set of regexes is huge, and even in the subset of regexes that are \emph{consistent} with the dataset, most are poor hypotheses (e.g., the regex \texttt{a|aa|aaa} for the dataset \texttt{\framebox{a {\posx} {\exsep} aa {\posx} {\exsep} aaa {\posx}}}).

We tackle these difficulties using a Bayesian approach -- assigning high prior probability to short regexes and high likelihood to regexes that explain the data well -- and performing approximate inference.
The Bayesian framework handles the issue of dataset uninformativity because uninformative examples lead to uncertainty in the posterior, which can be reported back to a end user (along with, say, a prompt to provide additional examples).
We deal with computational issues through a stochastic process recognition model that helps guide inference toward regions of high posterior probability by incrementally parsing example strings while generating the regex structure needed for parsing in a just-in-time fashion.
In Section 2, we describe related work.
In Section 3, we outline our high level approach.
In Section 4, we describe the model and implementation details.
In Section 5, we show that the method performs competitively with human learners in terms of recovering intended regexes from examples.

\section{Related work}

\textbf{Programming by example}. Related work has examined learning string transformations from input-output examples \cite{singh12} and inferring regular expressions for data extraction from examples of strings and extracted data \cite{bartoli16}.
Those methods learn programs that map strings onto strings, whereas we learn classifiers: programs that map from strings onto booleans.
These problems have a different structure; in some sense, we are working with less data, as boolean outputs have fewer bits of information than string outputs.

\textbf{Grammar induction}. We cast the problem of regex induction in terms of inferring a grammar, a topic that has been intensely studied in the field of grammar induction (GI) \cite{delaHiguera10}.
However, grammar induction research tends to focus on searching for the one true grammar that generated the data (e.g., MAP search as in \cite{stolcke94, chen95}), whereas we focus on approximating a Bayesian posterior over grammars.
In addition, grammar induction methods typically use \emph{merging} approaches, which initialize examples to a grammar that generates \emph{only} the positive examples, then searching through a space of local generalization transformations.
By contrast, our method incrementally builds a grammar using probabilistic programming techniques.
Finally, we handle character classes, which admit generalization from single characters to larger classes (e.g., \texttt{1} as a representative of the class of numeric digits), a feature not generally handled in grammar induction.

\textbf{Learning regexes from language}. Recent work has demonstrated good results in learning regexes from natural language \emph{descriptions} \cite{kushman13,locascio16}.
We focus on the problem of learning regexes from \emph{example strings}.
These approaches are complementary, as human users may find it more natural to describe some kinds of regexes using language (e.g., ``the substring \texttt{bob} must come after the substring \texttt{joe}'') and other kinds of regexes using examples, for instance when there is rich and possibly optional structure that is best shown with a list of cases, e.g., these examples for a number parser:
\begin{center}
  \texttt{\framebox{
      123 {\posx}
      {\vline height 2.5ex}
      123.456 {\posx}
      {\vline height 2.5ex}
      -123 {\posx}
      {\vline height 2.5ex}
      .456 {\posx}
      {\vline height 2.5ex}
      . {\negx}
      {\vline height 2.5ex}
      123.456.7 \negx
    }}
\end{center}

\textbf{Nonparametrics models of sequence data}. Our approach is inspired by nonparametric models of sequence data like the IHMM \cite{beal02} and the PDIA \cite{pfau10}.
However, those models learn representations that assign some positive probability to \emph{all} strings.
By contrast, we are interesed in learning classifiers, which accept some strings and reject others.


\section{Approach}

\subsection{Regex prior and likelihood}

A user has some target regex in mind and conveys it by generating a dataset of positive and negative examples (negative examples are optional, but very helpful).
For simplicity, we focus on a small but expressive set of regex features -- raw characters, disjunction, Kleene star, and character classes.\footnote{We use letters \texttt{[a-z]}, digits \texttt{[0-9]}, and the \texttt{.} wildcard, though it is straightforward to add more classes.}
We compute a posterior probability distribution on candidate regexes given the examples.
We use a prior favoring shorter regexes and a likelihood favoring regexes that explain the observed data well.
The prior is straightforward---we assign a prior probability that is exponential in the regex length and give higher weight to character classes than raw characters: $\prod_{r_i \in r} \gamma w(r_i)$, where $r_i$ is a regex token, $w(r_i) = 1$ for raw characters and $w(r_i) = \xi > 1$ for character classes, and $\gamma < 1$ controls the length prior (preference for shorter regexes).

The likelihood is less straightforward, as regexes are \emph{discriminative} objects, either accepting or rejecting strings.
The simplest idea, to define a binary likelihood indicating whether the regex accepts all positive examples and rejects all negative examples, leads to unsatisfactory results.
In particular, some strings are more \emph{representative} of some regexes than others.
For instance, the string \texttt{abababab} intuitively represents the regex \texttt{(ab)*} better than the regex \texttt{(a|b)*}.
To capture this intuition, we interpret regexes as generative models of strings, in particular as \emph{probabilistic regular grammars}.
A probabilistic regular grammar is a tuple $(N, S_0, T, R, P)$ containing a set of nonterminal symbols $N = \{S_0, ..., S_{n-1}\}$, a starting nonterminal $S_0 $, a set of of terminal symbols $T$, a set of rules $R$ for rewriting nonterminals, and a probability distribution $P$ on the rules for each nonterminal.
Rewrite rules take either the form $S_i \rightarrow t$ or the form $S_i \rightarrow t \, S_j$, where $t$ is a terminal.
We are primarily interested in distinguishing different grammatical structures rather than learning weighted distributions over rules, so we always take $P$ to be uniform.
Here is an example of a grammar representation of the regex \texttt{ab*b}:

\begin{center}
\begin{tabular}{l l}
$ S_0 \rightarrow \, a \, S_1$ & $p = \{1.0\}$\\
$ S_1 \rightarrow b \, | \, b \, S_1  $ & $p = \{0.5, 0.5\}$
\end{tabular}
\end{center}

There are two nonterminals, $S_0$ and $S_1$.
$S_0$ rewrites to $a \, S_1$ with probability 1, while $S_1$ rewrites to $b$ with probability 0.5 and $b \, S_1$ with probability 0.5.
We can sample strings by starting with the initial nonterminal $S_0$ and repeatedly sampling and applying rules until only terminals remain, e.g.,
$$S_0 \Rightarrow a \, S_1 \Rightarrow a \, b \, S_1 \Rightarrow a \, b \, b$$
This defines how we can sample strings given a grammar.
If we have a set of strings and want the likelihood that a particular grammar produced it, we can use probabilistic Earley parsing algorithm \cite{stolcke95}.

Probabilistic regular grammars clearly have three of the four regex features we desire---raw characters, disjunction, and Kleene star (recursive rules like $S_1 \rightarrow b \, S_1$), but they lack character classes.
We handle these by creating hard-coded character class nonterminals, e.g., $C_{\mathrm{ALPHA}} \rightarrow a \, | \, b \, | \, ... \, | \, z$, and also allowing rules of the form $S_i \rightarrow C$ and $S_i \rightarrow C \, S_j$.

Because we interpret regexes as grammars, it is much more convenient to do inference over grammars and then post-process this back to a distribution on regexes.
We next outline the inference approach.



\subsection{Inference: growing grammars}

Inference is challenging because most grammars are inconsistent with most example datasets--they either reject some positive examples or accept some negative examples.
Furthermore, even many consistent grammars will be implausible hypotheses.
To address this problem, we propose a recognition model approach that uses positive examples as a scaffold for incrementally building good grammars.
We start with an initial grammar that has only the starting nonterminal $S_0$ and no rules.
We then attempt to sample the positive example strings from our grammar.
If existing grammar structure suffices to generate a positive string, we reuse it with some probability; otherwise we sample new structure.

\begin{figure}[t]
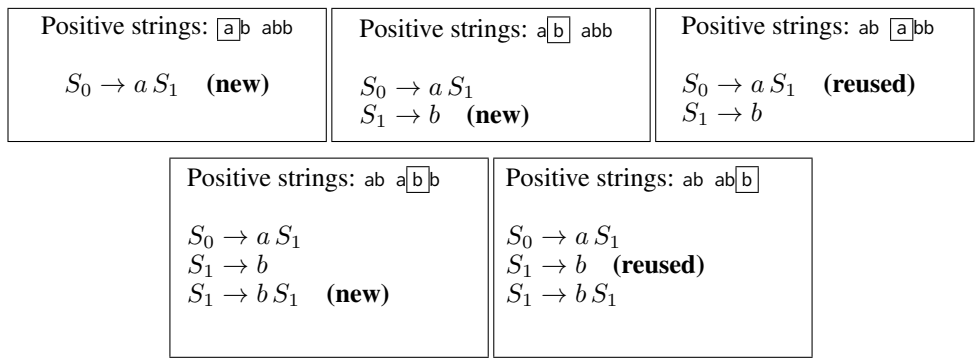

  \centering
  \framebox(120, 50)[t]{
    \parbox{0.24\columnwidth}{
      \vspace{1mm}
    Positive strings: \texttt{\fbox{a}b abb}\\
    $$S_0 \rightarrow a \, S_1 \quad \textbf{\textrm{(new)}}$$
  }
}  \framebox(120, 50)[t]{
  \parbox{0.25\columnwidth}{
    \vspace{1mm}
    Positive strings: \texttt{a\fbox{b} abb}\\ \\
    $S_0 \rightarrow a \, S_1$\\
    $S_1 \rightarrow b \quad \textbf{\textrm{(new)}}$
  }
}  \framebox(120, 50)[t]{
  \parbox{0.26\columnwidth}{
    \vspace{1mm}
    Positive strings: \texttt{ab \fbox{a}bb}\\ \\
    $S_0 \rightarrow a \, S_1 \quad \textbf{\textrm{(reused)}}$ \\
    $S_1 \rightarrow b$
}} \\

\vspace{2mm}
\framebox(120, 75)[t]{
  \parbox{0.27\columnwidth}{
    \vspace{1mm}
    Positive strings: \texttt{ab a\fbox{b}b}\\ \\
    $S_0 \rightarrow a \, S_1$\\
    $S_1 \rightarrow b$\\
    $S_1 \rightarrow b \, S_1 \quad \textbf{\textrm{(new)}}$
  }
}
\framebox(120, 75)[t]{
  \parbox{0.28\columnwidth}{
    \vspace{1mm}
    Positive strings: \texttt{ab ab\fbox{b}}\\ \\
    $S_0 \rightarrow a \, S_1$\\
    $S_1 \rightarrow b \quad \textbf{\textrm{(reused)}}$\\
    $S_1 \rightarrow b \, S_1$\\
  }
}

  \caption{Example of growing a grammar from positive strings}
  \label{fig:growing-grammar}
\end{figure}

To illustrate, we walk through a toy example (Figure~\ref{fig:growing-grammar}).
Assume the dataset \texttt{\framebox{ab {\posx} {\exsep} abb {\posx}}}.
We start with the first positive string \texttt{ab}.
We must rewrite $S_0$ to \texttt{a} and there are more characters after \texttt{a}, so we need a rule of the form $S_0 \rightarrow a \, X$.
The grammar contains no rules of this form (it's currently empty), so we must create a new one.
We sample the identity of $X$: with probability $\alpha_S$ we reuse an existing nonterminal at random and with probability $1 - \alpha_S$ we create a new nonterminal.
Assume we create a new nonterminal, $S_1$; then the grammar so far is $S_0 \rightarrow a \, S_1$.
We now must rewrite $S_1$ to the second character \texttt{b} and there are no more characters after \texttt{b}, so we need a rule of the form $S_1 \rightarrow b$; we add this, so our set of rules is now $\{S_0 \rightarrow a \, S_1, S_1 \rightarrow b\}$.

We now process the second positive string, \texttt{abb}.
We need a rule of the form $S_0 \rightarrow a \, X$.
There is an existing rule for doing this, $S_0 \rightarrow a \, S_1$.
We choose to reuse existing rules with probability $\alpha_R$ and create new rules with probability $1 - \alpha_R$.
Assume we choose to reuse the existing rule.
We then must rewrite $S_1$ using a rule of the form $S_1 \rightarrow b \, X$.
Assume we pick $X = S_1$.
Finally, we need to rewrite $S_1$ using a rule of the form $S_1 \rightarrow b$.
There is an existing rule of this form; again assume that we reuse the existing rule.
We have finished processing the positive examples.
If there were any negative examples, we would need to check that the grammar does \emph{not} derive any negative strings.
In this case, we have no negative strings, so we convert our grammar to a regex, resulting in \texttt{ab*b}.

The parameters $\alpha_R$ and $\alpha_N$ control preference for reusing existing rules and existing nonterminals, respectively.
This notation is suggestive of the \emph{concentration parameter} of the Chinese Restaurant Process.
Indeed, our reuse mechanism is inspired by nonparametric sequence prediction models like the IHMM \cite{beal02} and the PDIA \cite{pfau10}, which also place uncertainty over representation structure.
However, in those models, the probability of reuse depends \emph{dynamically} on the history of past reuse, whereas we use \emph{fixed} probabilities, which reduces computational complexity.
Additionally, because we end up post-processing the distribution on grammars to a distribution on regexes (which has its own prior and likelihood), $\alpha_R$ and $\alpha_N$ function as search parameters---they do not directly influence regex posterior probabilities but rather how we \emph{explore} the posterior.

\section{Method}

\begin{algorithm}[t]
  \DontPrintSemicolon
  \SetKwInOut{Input}{Input}
  \SetKwInOut{Output}{Output}
  \SetKwData{Scurr}{$S_{*}$}
  \SetKwFunction{sampleNT}{SampleNonterminal}
  \SetKwFunction{sampleRule}{SampleRule}
  \SetKwProg{myproc}{Procedure}{}{}

  \Input{Set of positive examples $X_{(+)}$, negative examples $X_{(-)}$, $\alpha_N$, $\alpha_R$, $\xi$}
  \Output{A sampled grammar}
  $N \gets \{S_0\}$; $R \gets \emptyset$\;
  \myproc{\sampleNT{}}{
    \eIf{flip($\alpha_N$)}{
      $N \gets N \cup {S_{|N|}}$; \KwRet new nonterminal $S_{|N|}$\;
    }{
      \KwRet a uniform sample from $N$\;
    }
  }

  \tcp*[h]{sample a rule of the form $S_i \rightarrow c$ or $S_i \rightarrow c \, S_j$ or $S_i \rightarrow C$ or $S_i \rightarrow C \, S_j$}\;
  \tcp*[h]{where $C$ is some character class containing $c$}\;
  \myproc{\sampleRule{$S_i$, c, size}}{
    $\mathbb{C} \gets$ the set of character classes that contain $c$\;
    sample $d \sim$ categorical distribution on $\{c\} \cup \mathbb{C}$\ with $p(c) \propto$ 1 and $c$ and $p(C_k) \propto \xi \, \, \forall C_k \in \mathbb{C}$\;
    \eIf{size = 1}{
      $r \gets$ new rule \fbox{$S_i \rightarrow d$}\;
    }{
      $S_j \gets $ \texttt{SampleNonterminal()}\;
      $r \gets$ new rule \fbox{$S_i \rightarrow d \, S_j$}\;
    }
    $R \gets R \cup \{r\}$\;
    \KwRet $r$\;
  }
  \ForEach{$x \in X_{+}$}{
    $S_{*} \gets S_0$\;
    \ForEach{character $c \in x$}{
      $R_* \leftarrow$ existing rules that rewrite $S_*$ \textrm{to} $c$\;
      \eIf{rules $\neq \emptyset$ and flip($\alpha_R$)}{
        $r \gets$ uniform sample from $R_*$\;
      }{
        $size \gets$ 1 \textbf{if} $c$ is final character in string \textbf{else} 2\;
        $r \gets $ \texttt{sampleRule(}$S_*, c, size$\texttt{)};
      }
      \lIf{$r$ uses a character class}{condition on sampling $c$ from that class}
      $S_* \gets$ the nonterminal on the right hand side of $r$\;
    }
    check that grammar doesn't parse any $x_- \in X_{(-)}$\;
  }
  \Return{(N, R)}
  \caption{Recognition model. \texttt{flip(p)} samples from a Bernoulli distribution with probability p.}
\end{algorithm}

We define the recognition model using probabilistic programming.
Probabilistic programming enforces a separation between defining generative models and performing inference on them: models are richly structured representations of probability distributions that can use conditionals, recursion, complex data structures, and external libraries in their definitions.
Inference algorithms are functions that operate on model programs to compute conditional distributions; languages built for probabilistic programming typically provide a number of built-in inference algorithms.
We use the probabilistic programming language WebPPL, which augments a mostly functional subset of Javascript with sampling, conditioning, and inference operators and provides a variety of inference algorithms (rejection sampling, enumeration, MCMC, SMC, and variational).
Although our actual implementation is in WebPPL, we present the generative model in pseudocode in Algorithm 1.
The WebPPL source code is available online at \texttt{http://github.com/<anonymized>}.

A benefit of expressing our recognition model as a probabilistic program is that we can flexibly change how we apply the model to the data.
For instance, in Algorithm 1 we processes positive strings in a fixed serial order, but it turns out to also be straightforward to process in random serial order, parallel order, or more complex ways.\footnote{For instance, we might perform inference on only the first positive string, use the discovered grammars as a starting point for inference on the second positive string, and so on.}
Also, having written our model as a generative program, the probabilistic programming language gives us immediate access to a variety of inference algorithms.
In simulations for this paper, we were able to quickly switch between rejection sampling, MCMC, bounded enumeration, and SMC by changing one line of configuration.
We found that SMC-based methods typically performed the best.

To summarize our pipeline, we run our recognition model on the dataset, which returns a sampled set of grammars.
We convert each grammar to a regex, compute the regex score (product of the regex prior and regex likelihood), and normalize scores to sum to 1.

\section{Experiment}

\subsection{Measure: $k$-best score}

We consider our method as a backend inference engine for a particular programming-by-example use case.
In particular, assume that the end user has a target regex in mind, but cannot write regexes herself (or wishes not to) but can easily generate examples and also \emph{recognize} the target regex from a ranked list of candidates.
Because human attention is limited, the user does not look at more than the top $k$ results in the list.
This corresponds to an inference algorithm both discovering the intended regex and assigning it one of the $k$ highest posterior probabilities.

We tested our method's ability to perform this target recovery problem using  example datasets generated by MTurk workers from the behavioral study of \cite{ouyang17}.
In that study, human subjects taught regular expressions by generating positive and/or negative examples for four simple regexes: \texttt{{\textbackslash}d{\textbackslash}d{\textbackslash}d{\textbackslash}d{\textbackslash}d}, \texttt{{\textbackslash}[.*]}, \texttt{.*s}, and \texttt{aaaa*}.
There were around 30 datasets generated per rule.
It was found that human teachers provided a small number of examples (between 1 and 13), used a mixture of positive and negative examples, and often  perturbed positive examples to generate informative minimal pairs.
For example, to teach \texttt{{\textbackslash}[.*]}, one teacher generated:
\begin{center}
  \texttt{\framebox{
      [dog] {\posx}
      {\vline height 2.5ex}
      dog {\negx}
      {\vline height 2.5ex}
      [cat] {\posx}
      {\vline height 2.5ex}
      cat {\posx}
      {\vline height 2.5ex}
      [123] {\negx}
      {\vline height 2.5ex}
      123 \negx
    }}
\end{center}
This kind of structure facilitates subsequent recovery from examples---human learners were often (but not always) able to recover the target regex from the fairly small example datasets.

We ran our algorithm on the human-generated datasets and evaluated the $k$-best score: the fraction of datasets for which the algorithm both discovers the intended regex and also assigns it one of the $k$ highest posterior probabilities (for $k= 1, 5, 10$).
Note that we do not necessarily expect scores near 1; \citeauthor{ouyang17} found that there was substantial variation in the ability of human learners to recover the target regex.
At one end of the spectrum, teachers provided quite helpful examples that all learners recovered the target from, while at the other end of the spectrum teachers provided unhelpful examples that no learners could recover the target from (e.g., one teacher generated just the single positive example \texttt{\framebox{ tjbuss \posx }} to teach the regex \texttt{.*s}).
As human-generated datasets vary in quality, we also break out the $k$-best scores by whether more than half of the human learners were able to guess the target regex.
Human ability to recover targets serves as a proxy for whether the target is theoretically recoverable from examples at all; we do not expect good performance when the targets are not recoverable in theory but we do desire good performance when the targets are recoverable in theory.
In principle, we would like to compare our algorithm's results to the true posterior, but if we could compute this we would have already solved our problem, so we compare to human data.

For these simulations, we used an alphabet of all characters that can be typed on a standard QWERTY keyboard.
We set $\gamma = 0.0002$ and $\xi = 10$.
Inside the recognition model we randomly sampled $\alpha_R \sim \{0.1, 0.2, \dots, 0.9\}$ and $\alpha_N \sim \{0.99, 1.0\}$.
We used an ensemble inference method that combined:

\begin{itemize}
\item 400 rejection samples (skipped if sampling rate was below 2 samples / second).
\item 5000 Metropolis-Hastings samples.
\item Particle Gibbs with serial processing of examples. We did multiple rounds: (1) 10 particles, 5 sweeps, 3s timeout, (2) 50 particles, 5 sweeps, 3s timeout, (3) 100 particles, 5 sweeps, 3s timeout, (4) 200 particles, 5 sweeps, 3s timeout, (5) 500 particles, 4 sweeps, 7s timeout.
\item Particle Gibbs with parallel processing of examples. We did multiple rounds: (1) 50 particles, 5 sweeps, 6s timeout, and (2) 100 particles, 5 sweeps, 6s timeout.
\end{itemize}

The particle Gibbs runs generally found the highest posterior regexes.
However, they also occasionally found grammars that correspond to regexes so long that they had negligible posterior probability, so we enforced timeouts.
Additionally, to further mitigate negligible regexes, we used the regex prior as an SMC importance distribution.


\subsection{Results}

\begin{figure}[t]
  \centering
  \includegraphics[width=0.7\columnwidth]{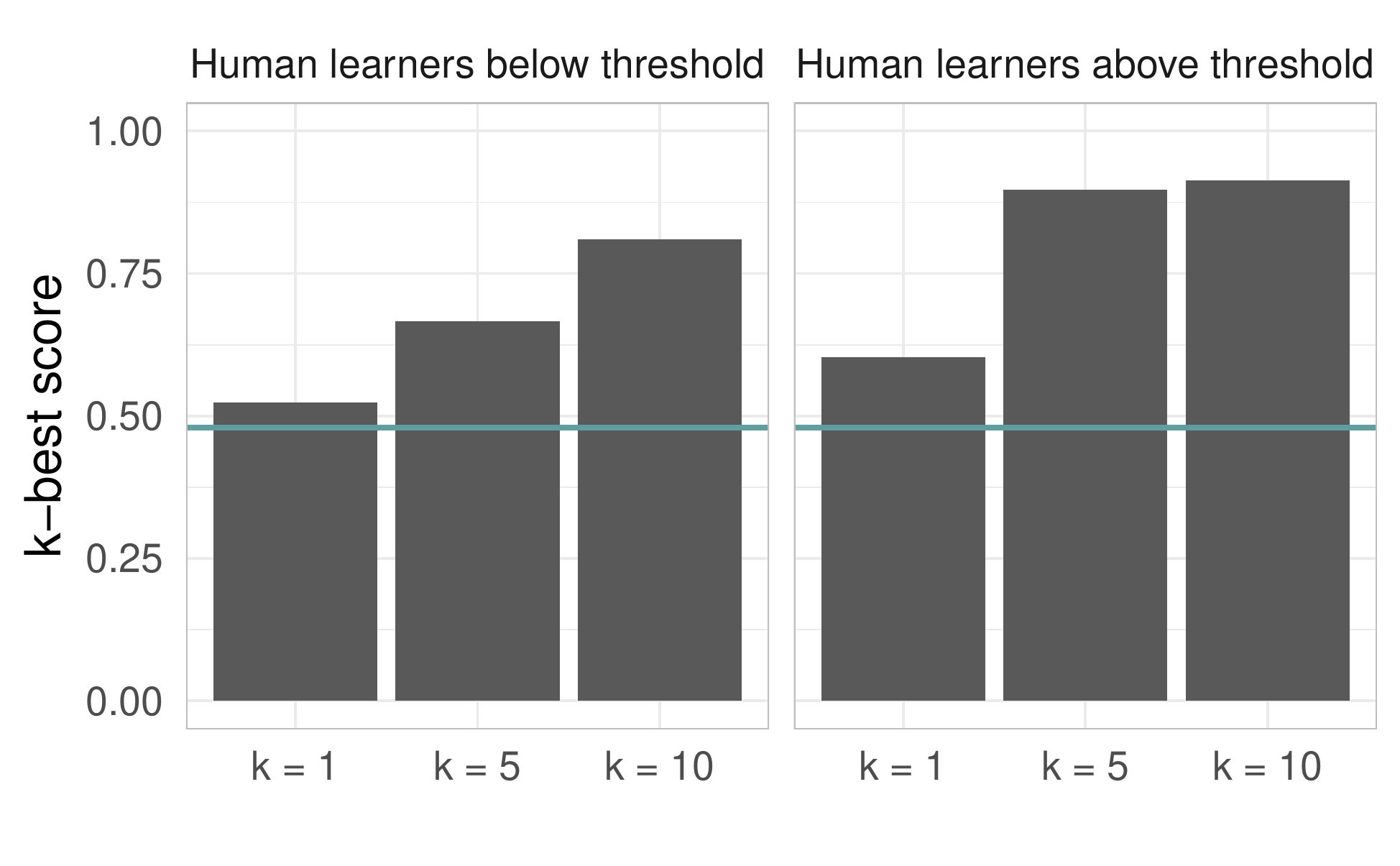}
  \caption{$k$-best scores for broken down by whether a majority of humans recovered the target regex or not. Blue horizontal line indicates the overall fraction of datasets where humans recovered the target.}
  \label{fig:result}
\end{figure}

See Figure~\ref{fig:result} for results.
Human learners recovered the target regex in 48\% of the datasets.
For all examined values of $k$, our algorithm had a score that was in some sense higher than that of humans.
Note that this is not an entirely direct comparison, as human subjects consider a richer class of rules for strings (the ``string has to be a mammal''), but it does suggest that a human teacher who provided examples to the algorithm could expect it to perform competitively with a panel of humans at recovering the target.
Indeed, for k = 10, the conditional probability of algorithm success given that more than 50\% of humans recovered the regex is high, at 92\%. In cases where where humans perform poorly (i.e., when fewer than 50\% of humans recovered the regex), the algorithm succeeds 80\% of the time.

We next detail some interesting comparison cases between the algorithm and humans.
First, consider a case where all humans failed to recover the target but the algorithm discovered the target and assigned it probability close to 1.
These 8 examples were used to teach the regex \texttt{.*s}:
\begin{center}
  \texttt{\framebox{fj38fj498js {\posx}
      {\exsep}
      r5iffffkkkks {\posx}
      {\exsepb}
      59yhkgyg7s {\posx}
      {\exsepb}
      FJEFJISFJs {\posx}}} \\
  \texttt{
    \framebox{SIJF\$\$FES {\negx}
      {\exsepb}
      48f94wfwh {\negx}
      {\exsepb}
      GRSRSRSFJh {\negx}
      {\exsepb}
      sw4wfJEHSFK {\negx}}}
\end{center}
Observe that the strings are fairly long and that the negative examples do not form minimal pairs with the positive examples.
Humans have limited memory and attention, so the long strings (and lack of easy comparison with related negative examples) likely explains why no human learner detected the pattern that strings must end in \texttt{s}.
By contrast, the algorithm does not have the same kind of memory and attention limitations, and it was able to recover the target in this case.
Conversely, here is a dataset where humans succeed but the algorithm fails.
These five examples were used to teach \texttt{\textbackslash[.*]}:
\begin{center}
  \texttt{\framebox{[hello] {\posx}
      {\exsep}
      hello] {\negx}
      {\exsepb}
      [hello {\negx}
      {\exsepb}
      []hello {\negx}
      {\exsepb}
      hello[] {\negx}
    }
    }
\end{center}
90\% of human learners recovered the target here but algorithm performs poorly.
According to the algorithm, the MAP hypothesis is \texttt{.hello]} with a probability of 0.25, whereas the target \texttt{\textbackslash[.*]} was ranked 31st with a probability of 0.001.
The MAP obviously avoids accepting the negative examples but it is obvious that the teacher cleverly used these negative examples: with a \emph{pedagogical} strategy.
The teacher carefully perturbed the positive example \emph{just} across the decision boundary a number of times to demonstrate crucial features of the rule.
Learners clearly understood this pedagogical strategy and succeeded because of it.
By contrast, the algorithm is not built on the presumption that example datasets will be generated in this helpful pedagogical manner, which yields a weaker and inaccurate inductive bias in this case.
To see this more clearly, we imagine that human learners would have no problem understanding these fictitious examples:
\begin{center}
  \texttt{\framebox{a {\negx}
      {\exsep}
      aaa {\negx}
      {\exsepb}
      aaaaa {\negx}
      {\exsepb}
      aaaaaaa {\negx}
    }
  }
\end{center}
But, as formulated so far, our algorithm cannot even run on such examples, as it requires positive examples to grow a grammar.
Handling pedagogically generated negative examples is therefore an important avenue for future work.

\section{Discussion}

This paper considered the challenging problem of programming regexes by example.
We focused on a small but realistic family of regexes with raw characters, disjunction, Kleene star, and character classes.
We developed a Bayesian inference approach, ``growing'' grammars using positive examples as a scaffold, and showed that this is both efficient and competitive with human performance on regex recovery.

There are three broad avenues for future work: (1) extending this technique to richer string languages, (2) making inference fast enough for real time use in an application, and (3) further improving the human factors aspects.

\textbf{Richer languages}.
Production regex systems handle a larger set of features.
It would improve the usefulness of our approach to handle features such as the optional modifier \texttt{?}, character ranges, and negation.
In addition, while there is existing work on learning regexes for string transformations or data extraction \cite{singh12,bartoli16}, it could be possible to do Bayesian approaches to these tasks by modifying the representations we learn.
Our method performs inference over probabilistic regular grammars, which are essentially equivalent to nondeterministc finite state automata; switching from \emph{automata} to \emph{transducers} would allow learning programs mapping strings to strings.
In a different vein, it could be useful to adapt our technique to learn larger class of languages from examples, such as context-free grammars (cf. \cite{bastani17}).
A starting point for this would be to switch from probabilistic regular grammars to Greibach normal form probabilistic context-free grammars.

\textbf{Optimizing inference}. Our software implementation often returned results quickly, in under 30 seconds, but sometimes took longer than would be acceptable for interactive use (e.g., up to 3 minutes).
The implementation has room for optimization but other inference algorithms should also be explored.
For SMC, particle Gibbs with ancestor sampling for probabilistic programming \cite{vandemeent15} could mitigate particle degeneracy.
Also, we could \emph{learn} to perform inference using a technique like neurally adapted SMC \cite{gu15}.

\textbf{Human factors}.
At the moment, our approach assumes human users can noiselessly generate examples from an intended regex and possibly recognize regexes from a list.
However, they can also label novel strings according to the target regex, suggesting that active learning could be worthwhile.
In addition, user sometimes carefully craft examples pedagogically; taking advantage of this could further improve inference quality.

\newpage



\bibliographystyle{ieeetr}
\bibliography{refs}

\end{document}